\documentclass[twoside]{article} 

\usepackage[none]{hyphenat}

\usepackage{times}				
\usepackage{url}                
\usepackage{scalefnt}
\usepackage{graphicx}			
\usepackage{amsmath}
\usepackage{relsize}
\usepackage{enumitem}

\usepackage{caption}
\usepackage{subcaption}

\usepackage{adjustbox}
\usepackage{array}

\newcolumntype{R}[2]{%
    >{\adjustbox{angle=#1,lap=\width-(#2)}\bgroup}%
    l%
    <{\egroup}%
}
\newcommand*\rot{\multicolumn{1}{R{25}{1em}}}

\usepackage[numbers]{natbib}	
\usepackage{lpscabs}			

\usepackage[pdftex,colorlinks=true,urlcolor=blue,citecolor=black,linkcolor=black]{hyperref}

\usepackage{array}
\newcolumntype{C}[1]{>{\centering\let\newline\\\arraybackslash\hspace{0pt}}m{#1}}

\usepackage[margin=0.99in]{geometry}

\begin{document}


\titlearea{Crater Detection via Convolutional Neural Networks}{
\\ Joseph Paul Cohen, Henry Z. Lo, Tingting Lu, and Wei Ding at the University of Massachusetts Boston (\{joecohen, henryzlo, ding\}@cs.umb.edu, lutingting@buaa.edu.cn)}

%

%



\subsubsection*{Introduction}
\vspace{-9pt}

Craters are among the most studied geomorphic features in the Solar System because they yield important information about the past and present geological processes and provide information about the relative ages of observed geologic formations.  We present a method for automatic crater detection using advanced machine learning to deal with the large amount of satellite imagery collected.

%
%

The challenge of automatically detecting craters comes from their is complex surface because their shape erodes over time to blend into the surface. Bandeira \cite{bandeira_automatic_2010} provided a seminal dataset that embodied this challenge that is still an unsolved pattern recognition problem to this day. There has been work to solve this challenge based on extracting shape \cite{urbach_automatic_2009} and contrast \cite{bandeira_automatic_2010,ding_sub-kilometer_2011} features and then applying classification models on those features. 


The limiting factor in this existing work is the use of hand crafted filters on the image such as Gabor or Sobel filters or Haar features. These hand crafted methods rely on domain knowledge to construct. We would like to learn the optimal filters and features based on training examples. In order to dynamically learn filters and features we look to Convolutional Neural Networks (CNNs) which have shown their dominance in computer vision \cite{krizhevsky_imagenet_2012}. The power of CNNs is that they can learn image filters which generate features for high accuracy classification.

CNNs are organized as a computation graph where the input image has computations performed on it and produce an output, then this output has computations performed on it, and this is repeated until an output layer which contains a prediction. There are many components to these networks but the most significant part to discuss is the convolutional layer and the fully connected layer.

\begin{figure}[h]
\begin{center}
\vspace{-10pt}
  \includegraphics[width=1\columnwidth]{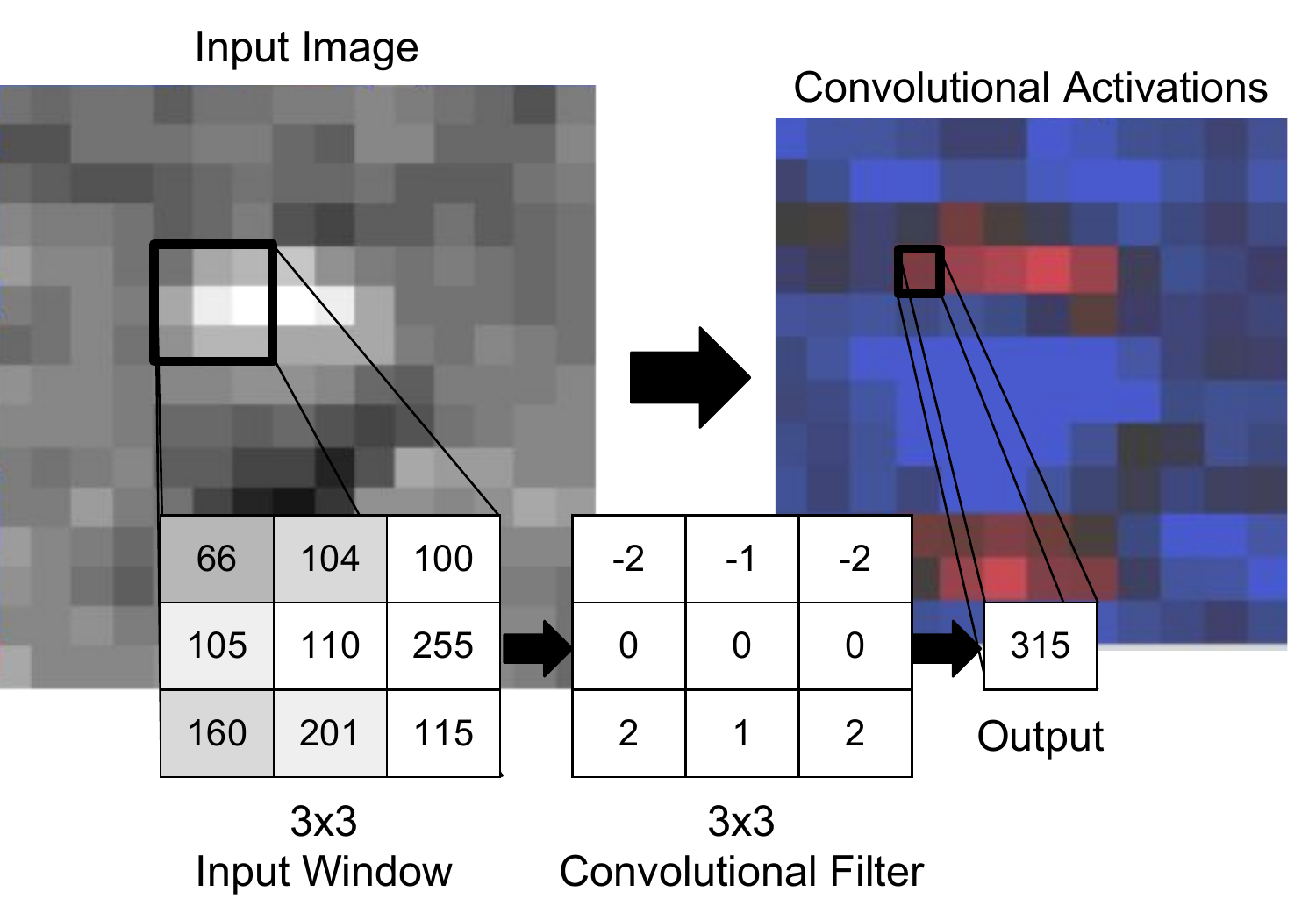}
  \caption[labelInTOC]{Convolutional layer}
  \label{fig:convlayer}
\end{center}
\vspace{-20pt}
\end{figure}

First we talk about a convolutional layer because this is the interface between the input image and the network. In Figure \ref{fig:convlayer} the process of a convolutional layer is shown. A sliding window is ran across the input image with a convolution applied at each position. Each pixel is multiplied by the corresponding filter value and summed together which results in a single value. A picture is formed from the results of these convolutions. This design is important because  spatial information is maintained to be used later in the network. In this example the filter values appear to be a sobel filter however the filters learned will be much different. During the learning process the values of the filters will be altered to minimize classification error of the entire network.

Next we talk about a fully connected layer. These layers have an input which is a vector $x$ of the previous layers outputs. These layers contain many nonlinear units. A nonlinear unit is a transformation $h_\theta(x)$ which produces a single value where $\theta$ is a vector of weights. The most modern practice is to use a rectified linear unit (ReLU): $h_\theta(x)=max(0,\sum_{i=0}\theta_i x_i)$. The $\theta$ value at this layer will be altered during training in order to minimize classification error.


\begin{figure}[h]
\begin{center}
\vspace{-10pt}
  \includegraphics[width=1\columnwidth]{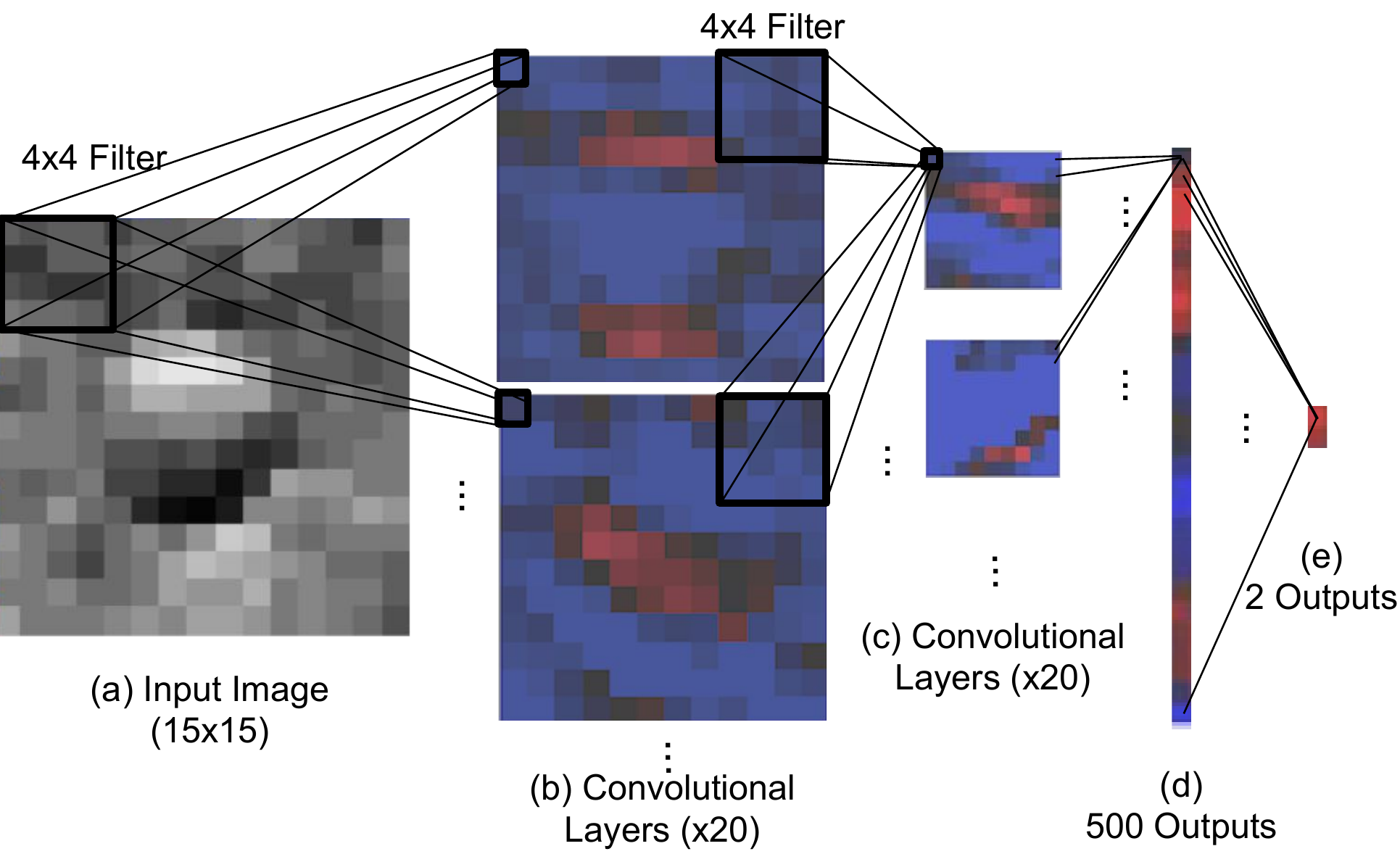}
  \vspace{-20pt}
  \caption[labelInTOC]{Crater Convolutional Neural Network (CNN) architecture computation graph. Each layer is identified with a letter and lines show processing from left to right.}
  \label{fig:cnnarch}
\end{center}
\vspace{-20pt}
\end{figure}

We combine the convolutional layers with fully connected layers in a specific configration that has achieved good performance on crater detection in Figure \ref{fig:cnnarch}. Layer $(a)$ is a 15x15 input image. Each candidate example is scaled to this size. Layers $(b)$ and $(c)$ are convolutional layers with 20 filters each of size 4x4. Each filter is passed over the filter in a sliding window fashion with a stride of 1 pixel. Unlike similar networks we don't use a pooling layer as it achieved poor results. Layer $(d)$ is a fully connected layer where each element of this layer is the result of $h_\theta(x)$ with unique $\theta$s for each element. Layer $(e)$ has just two outputs; each corresponding to a class. A softmax regression is used so that the results can be interpreted as a probability distribution between craters and non-craters.

The filter and $\theta$ values throughout the network are initialized randomly and incrementally modified using stochastic gradient descent to minimize classification error \cite{krizhevsky_imagenet_2012}. The term epoch is a training cycle that includes every training example. We omit the discussion on how training works for space.

\begin{figure}[t]
\begin{center}
  \includegraphics[width=1\columnwidth]{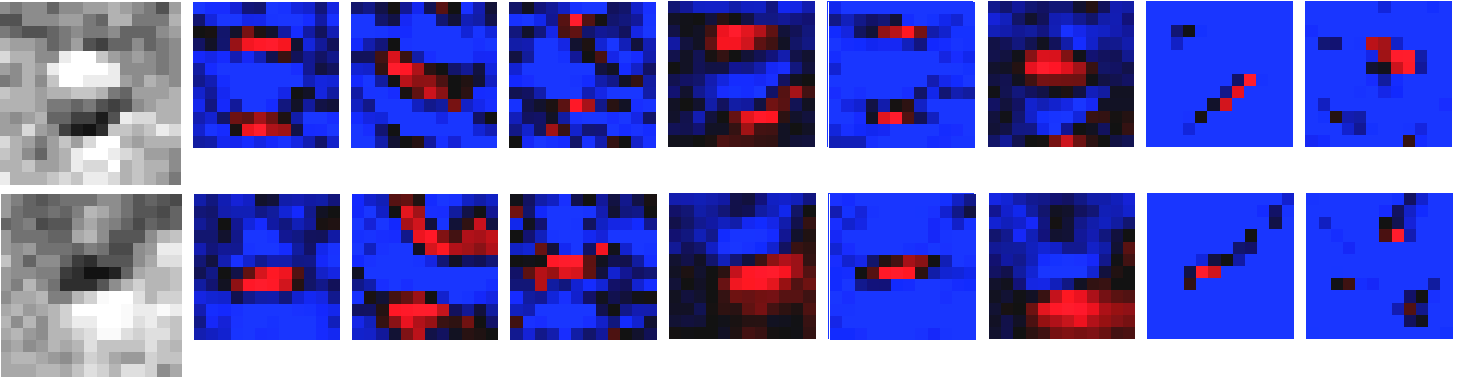}
  \vspace{-10pt}
  \caption[labelInTOC]{A crater and non-crater candidate are processed by the first convolutional layer. Eight filters with interesting activation patterns are shown to the right of each candidate image in false color.  Values are scaled and then colored to make these values visible and to maximize contrast within each square with blue = low and red = high.}
  \label{fig:filters}
  \vspace{-20pt}
\end{center}
\end{figure}

We can visualize the filters learned for craters to gain insight into what features are important to classification. We visualize 8/20 filters for layer $(b)$ on a crater and non-crater image in Figure \ref{fig:filters}. The filters, although crude, can be seen to show vertical edges as well as segmentation. The filters are learned based on what is nessary to minimize classification error on the training data so they do not directly relate to what a crater is but may be focusing on what is not a crater.

\vspace{-12pt}
\subsubsection*{Evaluation}
\vspace{-8pt}

We use Banderia's \cite{bandeira_automatic_2010} dataset\footnote{http://kdl.cs.umb.edu/w/datasets/craters/} to evaluate and compare our method.  It originates from the nadir panchromatic imagery footprint h0905\_0000 that was captured by the HRSC aboard the Mars Express spacecraft. This dataset is composed of 6 tiles (1700x1700 pixels each) with resolution 12.5m/pixel. A domain expert has labeled 3658 craters across all tiles. Banderia utilized Urbach and Stepinski's \cite{urbach_automatic_2009} highlight and shadow algorithm to avoid brute force sliding window candidate generation and reduce the total number of candidates that must be evaluated. This pre-processing step results in 2022 craters and 2888 non-craters across all tiles. The results of this pre-processing is what has been used in prior work and is used in our evaluation as well.

Prior work has used the F1-Score and cross-validation to evaluate their methods. The F1-Score is the harmonic mean between precision and recall: $F1 = \frac{2\cdot precision \cdot recall}{precision + recall}$. 10-Fold cross validation is a testing procedure where the dataset is randomly divided into 10 sections and then each section is selected as a testing set while the remainder is used as a trainng set with the resulting F1-Score averaged together.

Our results are shown in Table \ref{tab:stats}. The scores were obtained from the respective papers with the exception of ``Urbach '09'' which was obtained from \cite{bandeira_automatic_2010} where it is used as a baseline comparison. Our CNN approach is trained for 500 epochs during each fold evaluation. Our proposed CNN method obtains a higher score than all existing methods. We conclude that convolutional neural networks have a large potential to significantly change the way crater detection is performed. We believe further improvements to the design of the computation graph can increase performance much higher.

\begin{table}[t]
\begin{center}
\vspace{-10pt}
\begin{tabular}{|c|c|c|c|c|c|c|}

\multicolumn{1}{c}{Region}
&
\rot{Urbach `09} &
\rot{Bandeira `10}	& 
\rot{Ding `11}	& 
\rot{CNN}\\
\hline
\hline
West (1\_24+1\_25) 	& 67.89 	& 85.33 		& 83.98 	& 88.78  \\
\hline
Center (2\_24+2\_25)& 69.62 	& 79.35 		& 83.02 	& 88.81  \\
\hline
East (3\_24+3\_25)	& 79.77 	& 86.09 		& 89.51 	& 90.29  \\
\hline
\end{tabular}
\end{center}
\vspace{-14pt}
\caption{F1-score via 10-fold cross validation.}
\label{tab:stats}
\vspace{-10pt}
\end{table}

\captionsetup[subfigure]{labelformat=empty}
\begin{figure}[t]
    \centering
    \begin{subfigure}[t]{0.32\columnwidth}
        \centering
        \includegraphics[width=1\columnwidth]{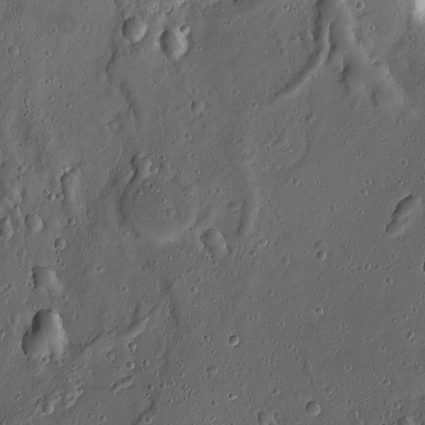}
        \caption{West 1\_24}
    \end{subfigure}
    \begin{subfigure}[t]{0.32\columnwidth}
        \centering
        \includegraphics[width=1\columnwidth]{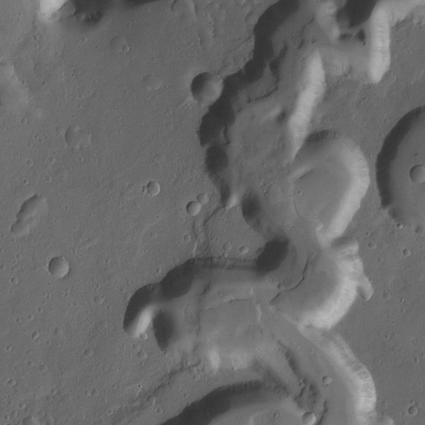}
        \caption{Center 2\_24}
    \end{subfigure}
    \begin{subfigure}[t]{0.32\columnwidth}
        \centering
        \includegraphics[width=1\columnwidth]{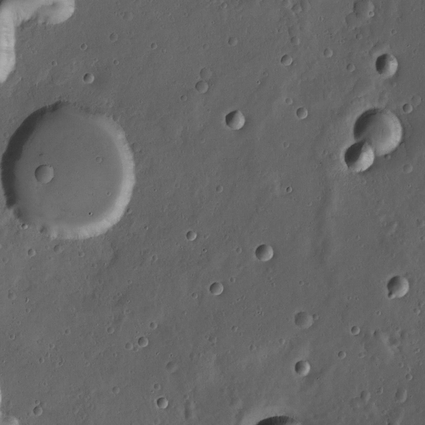}
        \caption{East 3\_24}
    \end{subfigure}
    
    \begin{subfigure}[t]{0.32\columnwidth}
        \centering
        \includegraphics[width=1\columnwidth]{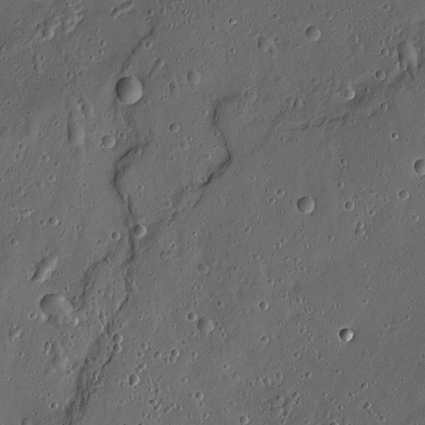}
        \caption{West 1\_25}
    \end{subfigure}
    \begin{subfigure}[t]{0.32\columnwidth}
        \centering
        \includegraphics[width=1\columnwidth]{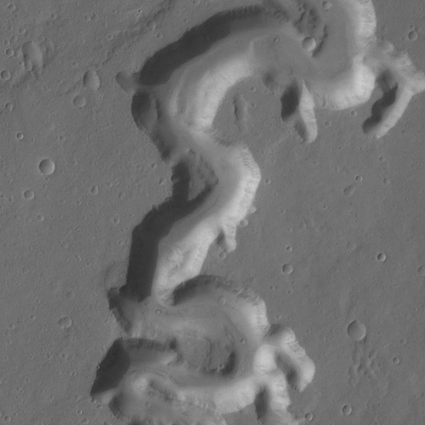}
        \caption{Center 2\_25}
    \end{subfigure}
    \begin{subfigure}[t]{0.32\columnwidth}
        \centering
        \includegraphics[width=1\columnwidth]{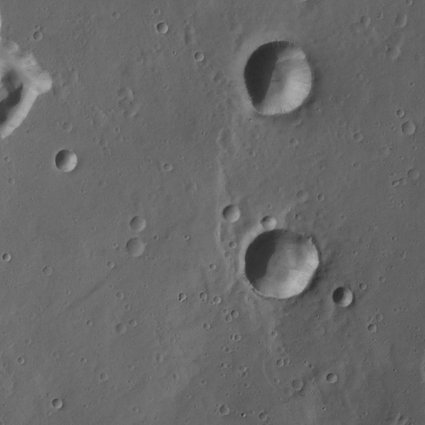}
        \caption{East 3\_25}
    \end{subfigure}
    
    \caption{Labels of each crater dataset tile.}
\end{figure}%

\bibliographystyle{IEEEtran}
\vspace{-12pt}
\subsubsection*{References}
\vspace{0pt}
\bibliography{joe,craters,neuralnetworks} 

\end{document}